\documentclass[11pt, a4paper, logo, twocolumn, copyright, nonumbering]{deepmind}

\usepackage[authoryear, sort&compress, round]{natbib}
\title{Diversity Through Exclusion (DTE): Niche Identification for Reinforcement Learning through Value-Decomposition}

\correspondingauthor{sunehag@deepmind.com}

\author[1]{Peter Sunehag}
\author[1]{Alexander Sasha Vezhnevets}
\author[1]{Edgar A. Du{\'e}{\~n}ez-Guzm{\'a}n}
\author[2]{Igor Mordatch}
\author[1]{Joel Z. Leibo}

\affil[1]{DeepMind}
\affil[2]{Google}

\begin{abstract}
Many environments contain numerous available niches of variable value, each associated with a different local optimum in the space of behaviors (policy space). In such situations it is often difficult to design a learning process capable of evading distraction by poor local optima long enough to stumble upon the best available niche. In this work we propose a generic reinforcement learning (RL) algorithm that performs better than baseline deep Q-learning algorithms in such environments with multiple variably-valued niches. The algorithm we propose consists of two parts: an agent architecture and a learning rule. The agent architecture contains multiple sub-policies. The learning rule is inspired by fitness sharing in evolutionary computation and applied in reinforcement learning using Value-Decomposition-Networks in a novel manner for a single-agent's internal population. It can concretely be understood as adding an extra loss term where one policy's experience is also used to update all the other policies in a manner that decreases their value estimates for the visited states. In particular, when one sub-policy visits a particular state frequently this decreases the value predicted for other sub-policies for going to that state. Further, we introduce an artificial chemistry inspired platform where it is easy to create tasks with multiple rewarding strategies utilizing different resources (i.e. multiple niches). We show that agents trained this way can escape poor-but-attractive local optima to instead converge to harder-to-discover higher value strategies in both the artificial chemistry environments and in simpler illustrative environments.
\end{abstract}

\begin{document}

\maketitle 

\section{Introduction}
The principle of competitive exclusion in ecology says that two species cannot inhabit the same niche, or put differently, cannot rely on the same set of limited resources \citep{volterra1928variations}. This is because if two species were to inhabit the same niche, one of them would have relatively more descendants than the other and thus over time come to dominate the niche, driving the other species to (local) extinction. Thus, speaking allegorically, we can say it is as if the prior presence of one species in a particular niche ``discourages'' others from joining it there. The repulsive effect that an occupied niche has on nearby species who might otherwise have occupied it is an important part of the reason why evolution on Earth produced a diversity of species \citep{levine2009importance}, not just a single generically optimal species like you might expect from a black box optimization algorithm.

Here we study how ecological concepts of competitive exclusion, niche discovery, and  diversity can be productively applied to a very different kind of adaptive system: a reinforcement learning (RL) agent operating in a complex environment. While evolution may be understood as a ``space-filling'' process, which over time generates a species for every as-yet-unfilled point in niche space \citep{schluter2000ecology}, in contrast, the standard image of an RL agent's learning process is the trajectory of a single point traveling through policy space by following a gradient and, after enough time, arriving at a globally optimal rest point \citep{sutton2018reinforcement}. In most of the RL literature, the central obstacle acknowledged to getting RL agents to discover radically new behaviors that a random policy would never emit by chance, is that of sparse reward and vanishing gradients. The idea is that gradients are often so small and inconsistent that no progress can be made. It's like learning is stuck in molasses far away from any local optimum\footnote{Indeed much of the recent work on exploration in RL has used the Atari game Montezuma's Revenge as the standard benchmark \citep{bellemare2013arcade}. In this game the molasses view really is accurate; a rather long sequence of actions must be emitted relatively precisely before the agent gets a single reward. As a result, gradients are often near zero and RL is very inefficient. Motivated by this, researchers developed a great many different methods for replacing sparse reward environments by dense reward environments with (hopefully) the same global optimum including a range of methods commonly interpreted as intrinsic motivations like curiosity (e.g.~\citep{pathak2017curiosity, burda2018exploration}).}. This is why the field has been mainly interested in models where an exploration bonus is added to the environment's default reward \citep{Singh04BC, badia20agent57}. The aim is to create new gradients where none would otherwise have existed (e.g.~\citep{eysenbach2018diversity}). In this paper, we explore how the ecological perspective may provide an alternative organizing metaphor that can be used productively to motivate a new reinforcement learning algorithm.

We can assume that a simulated species with an initially random configuration will usually evolve toward its closest possible niche. However, if that niche is already occupied then we expect the species to evolve in some other direction. In RL terms, unoccupied niches---which we may assume correspond to local optima---feature attracting gradients that would draw a nearby learner in. However, when a niche becomes occupied by another learner, then these gradients flip direction to instead point away from their location in policy space. An attractor becomes a repulsor. Thus greater initial diversity of occupied niches implies greater distance in policy space that a new learning agent joining the game must travel before finding still unexplored behaviors. Notice how this picture differs from that of sparse reward. Here, if many niches are already occupied, then a new agent may follow strong gradients all the way from its initialization point out to the frontier of unexplored behavior. In this view, the main hazard is premature convergence to local optima that are not good enough (poor niches), not vanishing gradients. Even bad local optima, if isolated from the better regions of policy space by vast intervening regions that are even worse, create basins of attraction that are difficult to escape once entered. The problem is not that gradients vanish, but rather that there are too many gradients, any one of which could all-too-easily suck in the agent and prevent it from learning more in the future. Once a niche has been occupied, then the attracting gradients diminish substantially, because the prior occupant would exclude the new entrant. With these considerations in mind, we can now articulate the central idea of this paper: instead of an additive exploration bonus aiming to create gradient where none exists, we propose to study a signal that modulates the impact of rewards on the learning that in effect is more like a multiplicative alteration of the rewards that can even set them to zero (or to a small value). It can be used to reduce or remove ``spuriously attractive'' gradients that would otherwise pull agents too strongly toward poor local optima and prevent their subsequent escape. Agents thus unencumbered are more free to explore, and therefore, able to discover parts of policy space they could not otherwise reach.

While our motivating analogies concern multi-agent systems, this paper is about an application of the competitive exclusion principle in the single-agent setting. It involves a population of sub-agents (policies) within a single agent. We are going to view this internal population as performing cooperative multi-agent reinforcement learning with value-decomposition networks \citep{sunehag2018value}. It will differ from multi-agent value decomposition because, in this case, we always know which sub-agent to credit with the whole return. Credit always flows to the policy that acted in the relevant episode and this fact impacts the resulting update rule.

To explore these questions, we introduce single-player RL environments with multiple niches, i.e., multiple different ways of earning reward, each associated with a different ``resource''. Each environment is a 2D world with dynamics driven by a programmable artificial chemistry (a reaction graph) \citep{banzhaf2015AC}. Different compounds can be used to achieve different rewarding reactions, and can sometimes be set up to run in sustainable (autocatalytic \citep{hordijk2012structure}) cycles. We include some simple and easy-to-discover ways of earning small amounts of reward that are only exceeded if an agent utilizes the more difficult strategies with substantial skill.  We show that the DQN control agent converges on the simple but less rewarding strategy, while our proposed algorithm avoids the low value strategy, converging instead to optimizing the high value strategy. Before this more complex task, we investigate the effect of the loss term / update our approach is based on, on illustrative tasks. For those, the natural control (ablation) is a multi-headed DQN that results if removing this extra loss / update. We see how our modification makes the agent more prone to seeking out different niches.

\section{Background} \label{sec:back}

\subsection{POMDPs, Dec-POMDPs, and Markov Games}

Reinforcement Learning (RL) \citep{sutton2018reinforcement} agents interact with an environment in cycles where the agent obtains rewards $r_t$, chooses actions $a_t \in \mathcal{A}$ and receives observations $o_t \in \mathcal{O}$. In a Partially Observed Markov Decision Process (POMDP), the observations and rewards are sampled based on an underlying state $s_t \in S$ that develops as a Markov Process where the distribution of $s_t$ depends only on $s_{t-1}$ and $a_{t-1}$. The agent is designed with the aim of achieving a high expected return $r_1+\gamma r_2+\gamma^2 r_3+  ...$ where $\gamma\in[0,1]$ is a discount factor. Suppressing the discount factor. we write that a POMDP is a tuple $(S, \mathcal{A}, \mathcal{O}, r)$.

Other settings for reinforcement learning are also possible besides POMDPs. For instance, in non-cooperative multi-agent reinforcement learning (MARL) there are multiple agents sending actions to the environment for each cycle and each receiving their own observations and rewards based on the environment's global state, which evolves according to a Markov Process where $s_t$ depends on $s_{t-1}$, and all the agents' individual actions in cycle $t-1$. This setting is a partially observed Markov game \citep{shapley1953stochastic, littman1994markov}. Here agents take actions based on an individual observation of the global state  $\mathcal{O} : S \times \{1, \dots , N\} \rightarrow \mathbb{R}^d$ and receive an individual reward $r^i: \mathcal{S} \times \mathcal{A}^1 \times \dots \times \mathcal{A}^N \rightarrow \mathbb{R}$. In each state, each player $i$ takes an action from its own action set $\mathcal{A}_i$. Formally, a partially observed Markov game is a tuple $(S, \{\mathcal{A}_i\}, \{\mathcal{O}_i\}, \{r_i\})$.

In cooperative MARL the setting is called a Dec-POMP \citep{oliehoek2016concise}. It lies midway between POMDPs and partially observed Markov games, combining features of both. It is like a Markov game in that it assumes a set of players, i.e., a vector of observations and actions for each global state. However, it is also like a POMDP in that there is only a single reward function which maps from the global state and the actions of each individual to a single scalar value. Formally, a Dec-POMDP is a tuple $(S, \{\mathcal{A}_i\},  \{\mathcal{O}_i\}, r)$.

\subsection{Reinforcement Learning Algorithms}

\subsubsection{Deep Q-Learning Networks for POMDPs}

A policy is a function $\pi$ that given a state $s$ produces an action $\pi(o(s))=a$. The value of this policy in state $s$ is $V^\pi(o(s))=\mathbb{E}\sum_{i=0}^\infty r_i\gamma^i$ where the expectation is with respect to the distribution generated by following $\pi$ after observing $o(s)$ in the relevant environment. $Q^\pi(o(s),a)$ is defined by the same expectation as $V^\pi$ but where the first action is $a$ instead of according to $\pi$. Further, with $Q^*=\max_\pi Q^\pi$ we have the optimal policy $\pi^*(o(s))=\text{argmax}_a Q^*(o(s),a)$. Some Reinforcement Learning algorithms aim to estimate $Q^*$ so as to chose actions with such an argmax. Q-learning, which is a temporal difference method, does so by minimizing the loss $\mathbb{E}_{o(s),a,o(s')} (Q(o(s),a)-r-\gamma \text{max}_{a'} Q(o(s'),a'))^2$. 

Deep Q-Learning Networks (DQN) \citep{mnih2015humanlevel} is a variant of Q-learning where $Q$ is represented by a deep neural network. It comes with a few differences to standard Q-learning to stabilize the method for this non-linear setting. Firstly, instead of directly performing gradient updates as samples comes in they are stored in a replay buffer from which they are sampled when updates are performed to make learning closer to i.i.d. Secondly, the term $\text{max}_{a'} Q(o(s'),a')$ is computed using a target network that is slowly updated so as to smoothly track the online network used to compute $Q(s,a)$. In all of the above, while we write the dependence as being on $o$ it might be on the whole history leading up to being in $s$. 

Note that the baseline DQN agent we use for the present work has been enhanced with several state-of-the-art generic improvements, namely multi-step lambda returns \citep{sutton2018reinforcement}, distributed actors adding experience into a joint replay buffer \citep{horgan2018distributedreplay} (though not prioritized replay), and a recurrent memory (an LSTM) \citep{hochreiter97lstm, KapturowskiOQMD19} to deal with partial observability. Strictly speaking it is no longer DQN, but it only has a subset of the additions of the R2D2 Atari agent \citep{KapturowskiOQMD19} which it most resembles, so that name would not be appropriate either.

\subsubsection{Competitive Exclusion in Partially Observed Markov Games}

In multi-agent situations where individuals have their own independent reward functions it is common to use the same algorithms as in single-agent reinforcement learning e.g.~\citep{leibo2017multiagent, perolat2017}. In this case each agent effectively views the others merely as being part of the environment.

Since agents have independent reward functions in this setting $r_1,\dots,r_N$ their incentives frequently come into conflict with one another. In the special case of two-player zero-sum rewards their incentives are fully conflicting: $r_1(s, a_1) = -r_2(s, a_2) ~ \forall s, a_1, a_2$. For one agent to win, its partner must lose. There is now a large literature exploring how self-play algorithms in two-player zero-sum settings can sometimes produce a natural curriculum in which agents continually learn new exploits and counters (e.g.~\citep{silver2018general, sukhbaatar2018intrinsic, leibo2019autocurricula, jaderberg2019human, baker2019emergent, vinyals2019grandmaster}). However, self-play does not always produce such open-ended exploration. Self-play may get stuck in a Nash equilibrium or limit cycle that doesn't require too much intelligence to implement, e.g.~consider rock-paper-scissors where all players in a Nash equilibrium choose a uniform random policy. Such policies are unlikely to generalize beyond their self-play setting, not even to matches against exploitable partners \citep{vezhnevets2020options, czarnecki2020real}.

\cite{leibo2019malthusian} study how partially competitive interactions can still provide impetus to explore when $N > 2$ and the game is not zero sum. They also developed a protocol where agents were trained in a multi-player mode but tested in a single-player mode. They showed that co-training in multi-agent mixed motive partial competition helped agents to escape poor local optima in their single-player task. In particular they introduced a task called Clamity in which a player, imagined as a larval clam, has an action to settle at its current location. Whenever it settles it starts building a clam shell around itself, causing a stream of reward to begin, which last until the end of the episode and is proportional to the size of the clam they manage to build. If too many players settle too close to one another their clams do not have enough space to grow and end up smaller than they would otherwise be. Thus training with many players forces the agents to learn to spread out through space before settling, and this leads them to discovering that they can earn even more reward by settling in nutrient rich patches, far from their starting location.

\subsubsection{Value Decomposition for Cooperative MARL in Dec-POMDPs}

In the cooperative setting where all rewards coincide and, thereby, being a team reward, a special class of algorithms have been developed. For this article, the Value-Decomposition-Networks (VDN) \citep{sunehag2018value} approach is relevant for the derivation of our new single-agent algorithm. While a larger literature has developed over recent years building further on this idea, e.g. \citep{rashid2018qmix}, we here rely on the basic version from \citep{sunehag2018value}. The approach learns a joint $Q$-function of the form $Q((o_{1}(s),...,o_{N}(s)),(a_1,...,a_N))=\sum_i Q(o_{i}(s), a_{i})$ where $a_i$ and $o_{i}(s)$ are individual observations and actions for agent $i$. Note that the joint argmax can be calculated by individual maximization. The Q-learning loss on the joint leads to a gradient that propagates down through each individual in a simple and elegant manner.

In cooperative MARL, VDN can be understood as learning to assign credit for achieving global return to individual agents in a way that depends on how well their state and action can predict the global return.

\section{Single-Player Niche Identification}\label{sec:niche}

We consider environments with multiple separate rewarding strategies that are not refinements of each other but rather are such that utilizing one makes it harder to find the others. We do not want our agents to perform very large amounts of random or novelty seeking actions, but want instead to consider the more realistic setting where a smaller amount of such exploration must be used to rapidly move toward new niches/optima by following a gradient. We take inspiration from the principle of competitive exclusion to derive a multi-agent inspired single-agent approaches to niche identification.

We utilize the idea of Value-Decomposition-Networks \citep{sunehag2018value} from cooperative multi-agent reinforcement learning but adapted here to the case where there is only only player in the environment at a time. We would like an agent that consistently utilizes a particular rewarding strategy to have a higher value estimate for that strategy than an agent that less frequently uses it, and to see diminishing value when others increase their frequency of using it.

\subsection{Diversity Through Exclusion (DTE)}

We consider a single agent as though it contains multiple cooperating agents. Thus it makes sense to view it through the Dec-POMDP framework of cooperative MARL and rely on the approach of Value-Decomposition-Networks (VDN) \citep{sunehag2018value}, though here the implications of the VDN approach are different from its original setting. Since all the ``sub-agents'' are inside one agent, and only one can act at a time. Credit assignment should flow only to the agent who actually acted. Please note, from here to the end of the paper we will be assuming for simplicity that all rewards are non-negative.

\begin{figure}
    	\includegraphics[width=7cm]{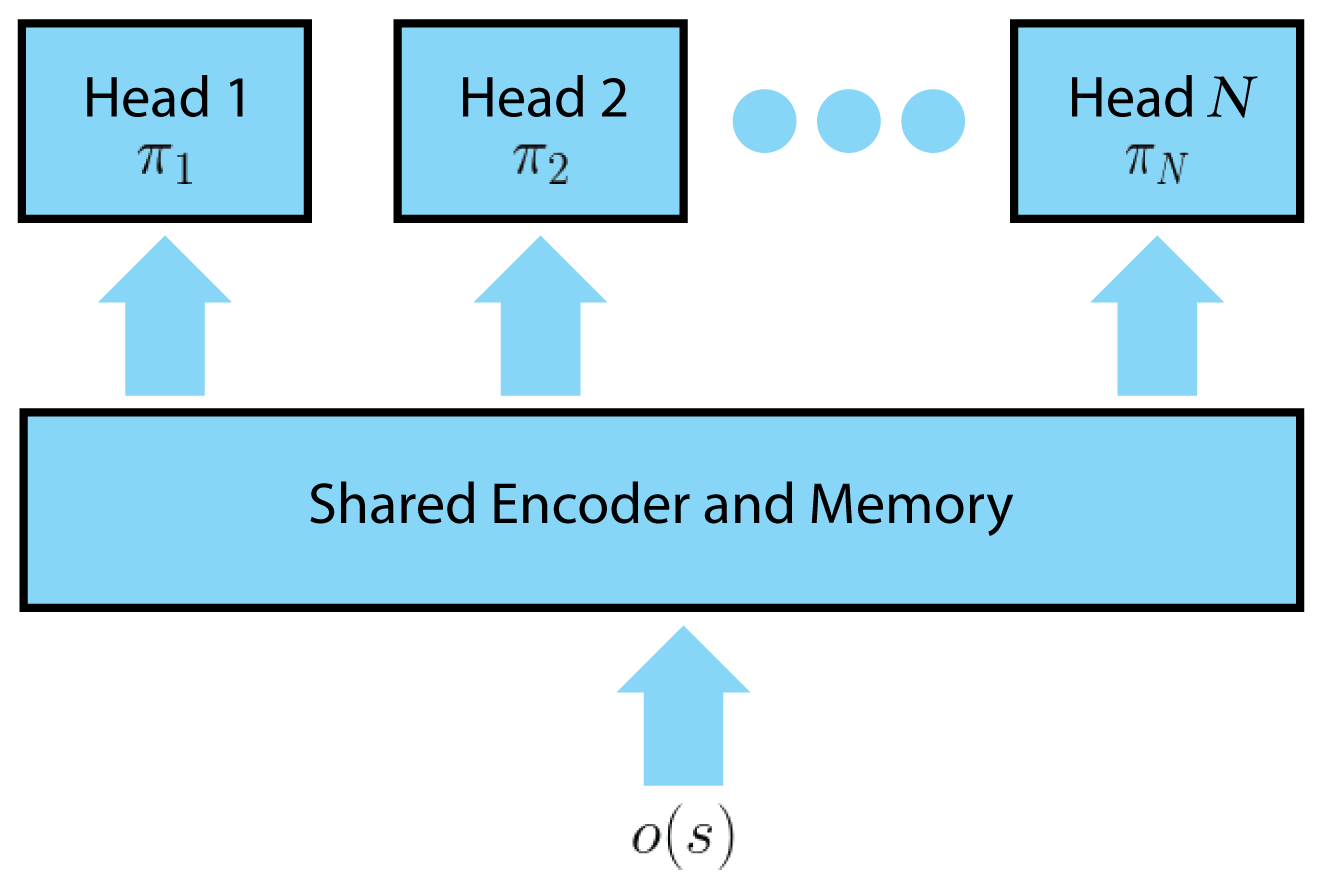}
    	\caption{Multi-head network architecture with multiple policy heads from a shared memory (LSTM) and encoder (a two layer convolutional network)}
        \label{fig:multihead}
\end{figure}

Our approach is based on sampling a policy $\pi_i$ from a population for each episode and using it to generate an experience trajectory. These policies will, in our experiments, be defined by $N$ different policy heads on top of a common core encoder network as in Fig.~\ref{fig:multihead} and similar to Bootstrapped DQN \citep{osband2016boot}, which learns with DQN updates for each head (as our baseline will for their own generated experience) based on all experience and acts in testing based on a majority vote.

We can convert any POMDP into a Dec-POMDP with multiple cooperating sub-agents inside a single agent by the following construction. Let there be $i = 1, \dots, N$ sub-agents. Since a POMDP has only one observation function $o: S \rightarrow \mathbb{R}^d$, and we consider all $N$ sub-agents to inhabit the same body, we define all $N$ observation functions of the Dec-POMDP to be the same as one another: $o_{i} = o~\text{for}~i=1,\dots,N$. A Dec-POMDP has a single scalar reward function $r: S \times \vec{\mathcal{A}} \rightarrow \mathbb{R}$ which depends on the state of the environment $s \in S$ and the joint action vector $\vec{a} \in \vec{\mathcal{A}}$. At the start of each episode, sample a particular sub-policy $\pi_i$. Let $\vec{a}(s)=a_{i}$ where $a_{i} \sim \pi_i(o(s))$ is the action taken by the $i$-th sub-agent in response to the observation $o(s)$. 

Now using this multi-player Dec-POMDP representation of the single-player POMDP, we can straightforwardly apply the value decomposition networks approach \citep{sunehag2018value}. Let the global state-action value function $Q$ be decomposed as 
$$Q(o(s), a)=\sum_{i=1}^{N} Q_i(o_i(s), a_i)$$
where $Q_i$ is the state-action value function of policy $i$ but here $o_i(s)=o(s)$ and $a_i=a$ since all sub-agents experience the same observations and actions. 

In our case, we have more information than in cooperative MARL. For each state $s$ we know which sub-agent $j$ was responsible for selecting the joint action $\vec{a} = a_j(o(s))$. Instead of learning by backpropagation how to assign credit to agents for achieving rewards, we can directly use our true knowledge of which sub-agent was responsible for each action taken and reward obtained to assign credit appropriately from the very start of training. This amounts to updating the responsible sub-agent toward a target using the true sequence of rewards while simultaneously updating all other sub-agents with the same observation and action sequence but all rewards replaced by $0$.

Thus for every trajectory $(o(s),a)$ and policy $i$, a learning update is performed where a gradient step is taken for the loss
$L(o(s),a, i, \{Q_j\}_j, R) =$ $$(Q_i(o(s),a)-R)^2 + (\sum_{j\neq i}Q_j(o(s),a))^2,$$
which by the triangle inequality is not smaller than the standard VDN loss $(\sum_{j}Q_j(o(s),a)-R)^2$.

The reasoning that motivated this update was to have a player's average return from a niche be proportional to how much of its utilization it represents. 
As one sub-agent increases the frequency with which it visits a particular state associated with a poor local optimum it should decrease the effective rewards experienced by other sub-agents who start later on to visit that local optimum, in effect, ``claiming'' its associated niche and competitively excluding the others. Other sub-agents who would have stumbled upon the same local optimum later in their training then, once they find it, experience lower than usual rewards there, potentially helping them to avoid distraction (premature convergence) and continue to explore.

The nearest baseline to compare with this approach is a multi-headed DQN and doing so in detail elucidates the effect of the extra loss term / updates. In the baseline we merely train a population of sub-agents using the experience they generated while in control of the single player. Whereas, in our Diversity Through Exclusion (DTE) algorithm, we also update across policies where the experienced rewards of all other policies besides the one acting are all replaced by $0$. This has the effect of decreasing the value estimates of experienced states for sub-agents that do not themselves generate trajectories involving those states.

\subsection{Interpretation as Multiplicative Gating}

Having derived the DTE loss function and updates, we now discuss another important way of intuitively understanding its actions. Note however, that in particular for non-linear value functions, this interpretation is mathematically just a loose approximation to provide intuition. Note that, if the state information told the agent which sub-agent had acted in a given state, then minimizing the loss would mean aiming for $Q_j(o(s),a)=0$ if $s$ says that $j$ did not act and for $Q_j(o(s),a)=R$ if $s$ did act and $R$ is the return the environment is expected to produce. In reality, $s$ might contain some information about what head acted and then we can loosely have
$$Q_j(o(s),a)\approx Pr(j|o(s),a)R$$
where $Pr(j|o(s),a)$ is an implicit prediction of acting head. This view of it makes the method related to empowerment methods such as DIAYN \citep{Eysenbach2019DIAYN}, VIC \citep{Gregor2017VariationalIC} and DADS \citep{Sharma2020Dynamics-Aware}, though our method is entirely reward driven and those are without environment rewards and give rewards to a policy for being identifiable, i.e. for a policy being well discriminated from the other policies leading to locomotion arising in continuous control as it reaches more different states. For our method DTE, where the representation and task allows the agents enough possibilities for making heads distinguishable from each other based on generated trajectories, and for the value functions to implicitly learn to do this well, the exclusion will gradually decrease. This is not a bad thing for our purposes, as the exclusion lets agents discover different niches early on while the later identifiability allows more heads to occupy the most rewarding part of the state space and learn to optimize that niche. We indeed see this in both of our artificial chemistry domains, where eventually all heads learn the best strategy. It should, however, be noted that even if value estimates can get higher from distinguishing heads, unlike empowerment methods, DTE does not reward the policies for being well discriminated from each other.

\section{Experimental Evaluation} \label{sec:exps}

We define two kinds of tasks for our experimental evaluation. One of them is a maze which we use for simpler tests where the outcome is easily understood. The other kind of task is based on an artificial chemistry platform where we are able to create complex compositional tasks involving several different resources, giving rise to multiple niches.

\begin{figure}
    	\includegraphics[width=0.97\linewidth]{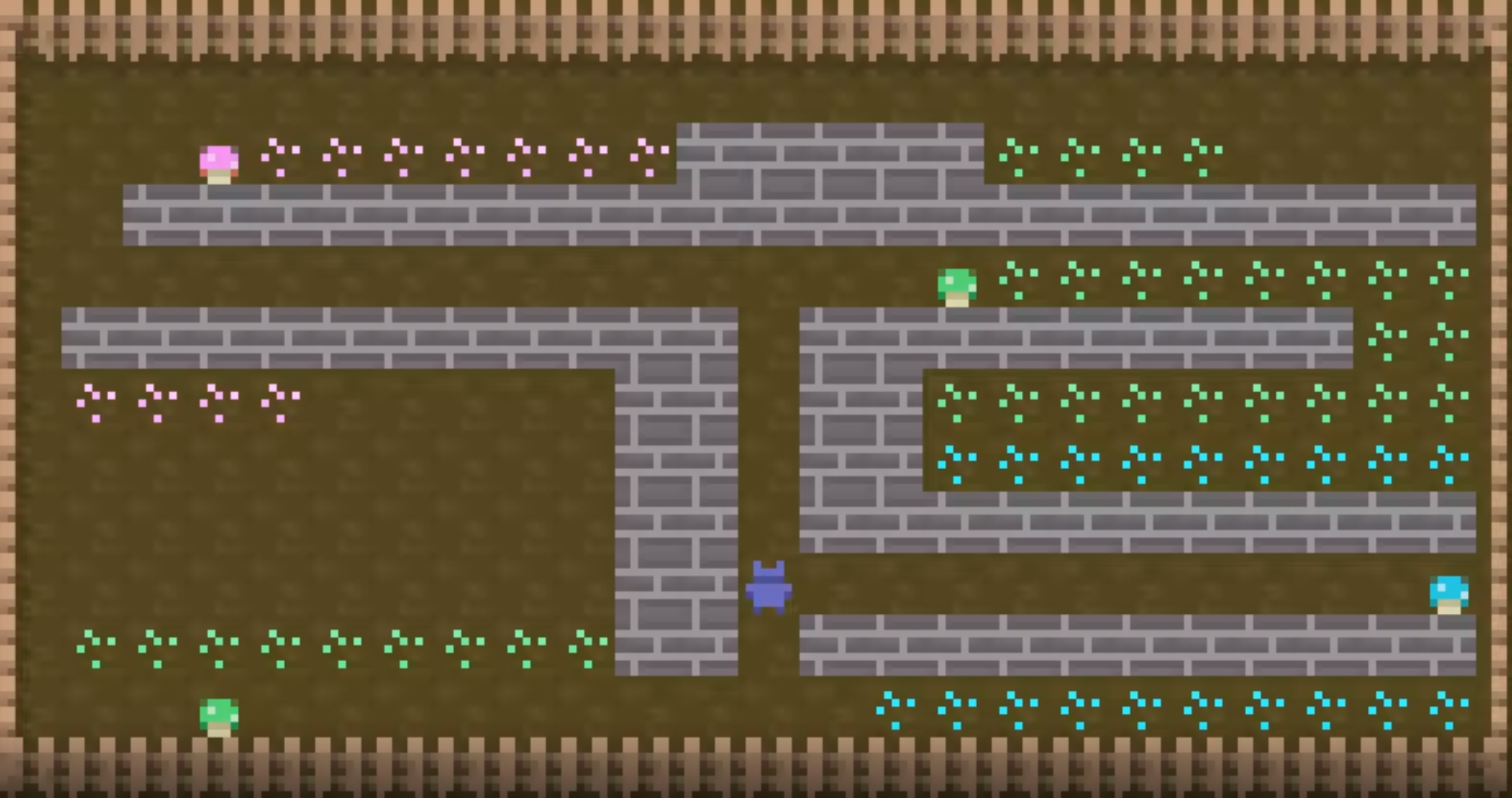}
    	\caption{The maze environment.}
    	\label{fig:maze}
\end{figure}

\subsection{Maze}
The maze is based on a 2D map (Fig.~\ref{fig:maze}) with mushrooms of different colors that give different rewards as they are digested over $25$ time steps. Digestion starts when the agent reaches any mushroom and the episode ends when digestion finishes. Agents cannot move during digestion, so only one mushroom can be consumed per episode. The maps also features walls (that agents cannot walk through) and decorations (flowers that agents can walk through) in the same colors as the mushrooms. The actions are up, down, left, right and the observations are a square around the agent.

As can be seen in Fig.~\ref{fig:maze}, the most rewarding mushroom, the red one which gives $1.25$ per step, is tucked away more in the top left corner. The agent spawns at the bottom of a corridor. The least rewarding mushroom, the green one, which gives $0.75$ per step, is easily accessed to the right from the top of that corridor and there is another straight down and to the left. The blue mushroom, which gives $1.0$ per step, is located straight to the right. As we will see in our experiments, an agent is very likely to first learn a path to either the green or the blue mushroom. The question is if the agent will then afterwards learn to switch to the most rewarding red mushroom. As we equip our DQN-style agents with only a constant $\varepsilon=0.1$ amount of intrinsic randomness, this is less likely (as most of its non-random steps will be directed elsewhere).

We run both DTE and the baseline DQN agent with $9$ heads. The agent's neural network consists of a two-layer convolutional network with output channels of $16$ and $32$, kernel shapes of $8$ and $4$, and strides of $8$ and $1$. These numbers were chosen to line up low level filters precisely with the environment's $8 \times 8$ sprites\footnote{These convolutional network parameters are standard for work using Melting-Pot-derived environments (see \citep{leibo2021scalable}).}. The convolutional network feeds into an LSTM with $128$ hidden units and then to the $9$ heads that are two-layer MLPs with $128$ hidden units and outputs the predicted values for each action. The acting head is sampled uniformly for each episode. For the agent we refer to as DQN, we perform the normal DQN update (as in the DQN variant described above) for the head that has acted in the episode. For DTE, we also perform the additional update for the other heads where a gradient step is taken for loss $(\sum_{j\neq i}Q_j(s,a))^2$, i.e. decreasing the value (we assume non-negative rewards) for the other heads in this state. 

\begin{figure}
    	\includegraphics[width=0.9\linewidth]{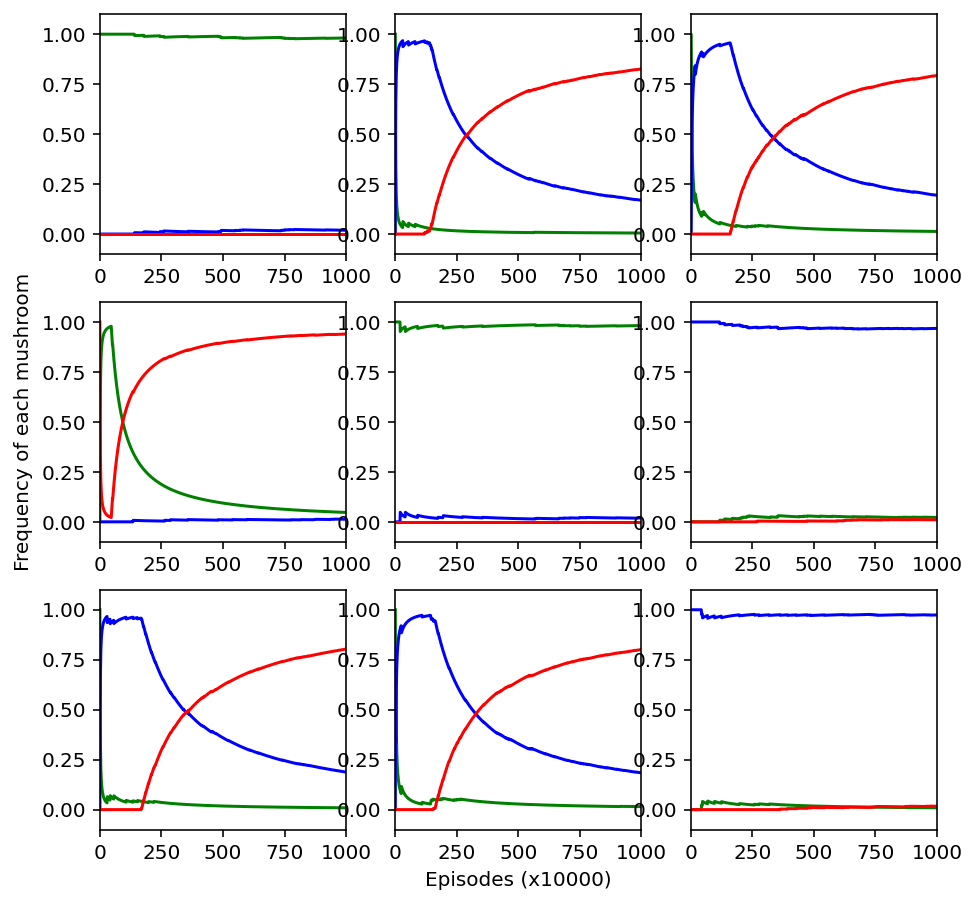}
    	\caption{Results for experiment with DTE with 9 heads in the Maze task}
        \label{fig:maze_dte}
\end{figure}

\begin{figure}
    	\includegraphics[width=0.9\linewidth]{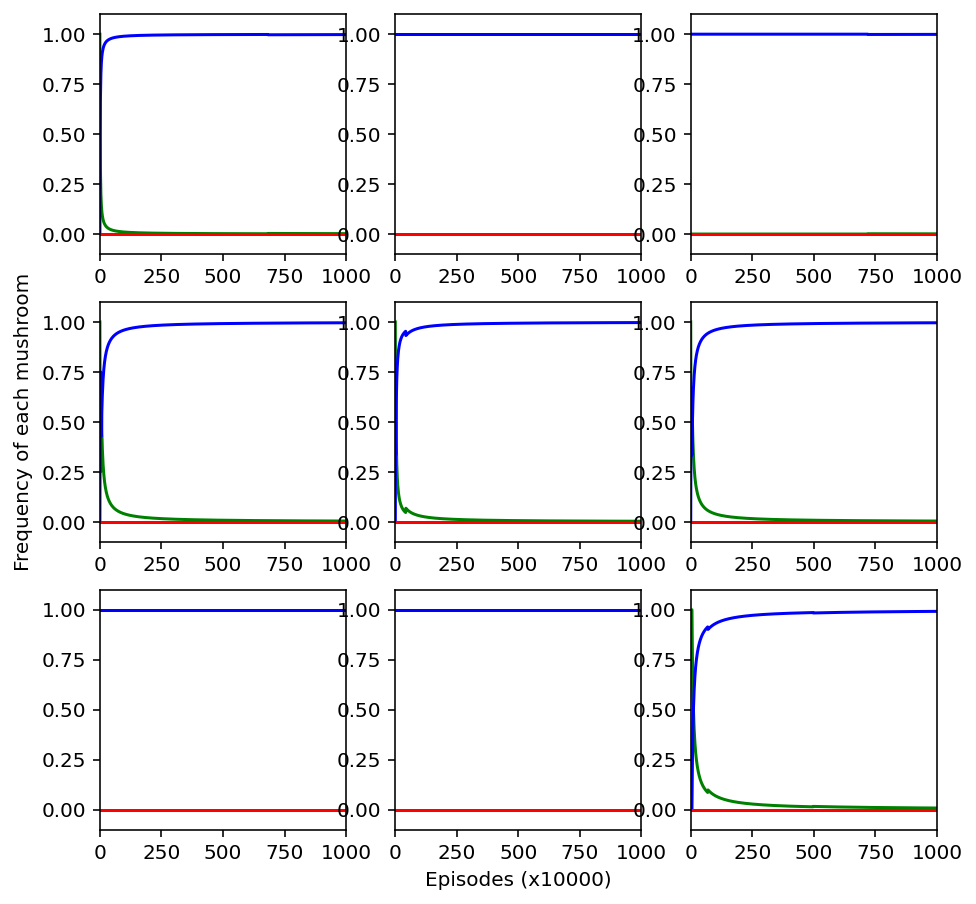}
    	\caption{Results for experiment for DQN with 9 independently updated heads in the Maze task}
        \label{fig:maze_dqn}
\end{figure}

We display the results for all $9$ heads of DQN in Fig.~\ref{fig:maze_dqn} and DTE in Fig.~\ref{fig:maze_dte}. For both agents, we can see how all $9$ heads start out by quickly learning a path to either a green or a blue mushroom. For DQN, all heads settle very quickly on the blue mushroom straight to the right (the five that started with green switch very fast to blue since that is very simple to find and gives more reward). Meanwhile, for DTE we see how two heads remain on green, two stay with blue, while the rest turns to red. Hence, DTE has produced a set of sub-policies that collectively identified all the niches. If one wants a test policy achieving the highest reward then one can simply pick the recently best performing head. DTE exhausted the niches, while the $9$-headed DQN did not. The two agents have identical network architectures and the only difference between the two is the extra cross-head update. Thus the results on this simple maze task illustrate well the difference it makes.

\subsection{An Artificial Chemistry Platform}

To enable the creation of rich environments with many completely different rewarding strategies separated from one another by local optima, we create a platform for compositional environment design inspired by artificial chemistry, where environments are defined by reaction graphs \citep{banzhaf2015AC}. Our specific environments are inspired by work on autocatalytic sets (RAF sets) \citep{hordijk2012structure} and the notion of a chemical reaction system used in the RAF sets literature. The most interesting environment we consider here used here involved metabolic cycles aimed at being what \citep{hordijk2012structure} calls Reflexively Autocatalytic and Food generating (RAF). The notion of \citep{hordijk2012structure}, that autocatalytic systems have such RAF subsystems, has critically guided our task design. An important property of this class of environments we introduce is that they are compositional, hence a more complex task world can be created by merging two simpler ones and, thereby, featuring the rewarding strategies of each as well as potentially new strategies. 

Reactions occur stochastically when reactants are brought near one another and are defined by a reaction graph (see Fig.~\ref{fig:chemistry_metabolic}) showing reactants (the required molecules) going into a reaction node and products at the end of outgoing edges. Tasks are defined by a reaction graph together with reward specification and a map setting initial molecules.  Agents can carry a single molecule around the map with them at a time. Agents are rewarded when a specific reaction---such as metabolizing food---occurs with the molecule in their inventory participating as either a reactant or a product. Each reaction has a number of reactants and products, occurs at different rates that can depend on whether or not it is in an agent's inventory. As an example, metabolizing food in the metabolic cycles environment has a much higher rate in the inventory where it generates a reward versus outside where it represents the food molecule dissipating.  

\subsubsection{A small artificial chemistry task:}

\begin{figure}
    	\includegraphics[width=6cm]{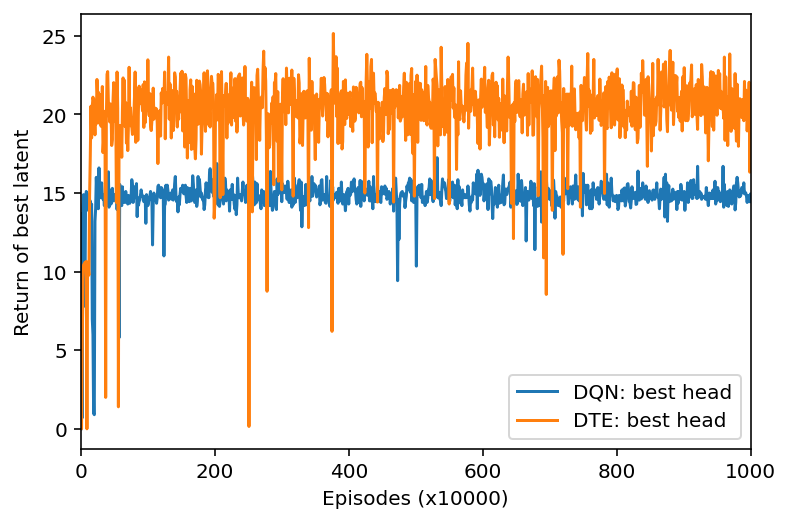}
    	\caption{Best heads (average of all episodes) of $9$-headed DQN vs DTE on Simple Artificial Chemistry}
        \label{fig:simple_chemistry_best}
\end{figure}

This first task (reaction graph in Fig.~\ref{fig:simple_chemistry_best}) only features two kinds of molecules, red and green. They are both in a rewarding identity reaction where the molecule is both the reactant and product. While all reactions in this task has the same rate, the red identity reaction gives reward $0.1$ and green only $0.75$. However, both are also involved in a pair identity reaction where two red are both reactant and product or two green react but remains two green. The two green gives $0.25$ while two red only $0.15$. Hence, while only having learnt that holding one molecule in the inventory is rewarding, it is best to hold the red one but when the pair reactions are discovered it is better to have green. However, an agent might first learn to consistently go to red and then later discover that moving that red over to the other gives more, while never discovering that bringing the two green molecules together is even better.

\begin{figure}
        \hspace*{-0.5cm}
    	\includegraphics[height=0.9\linewidth]{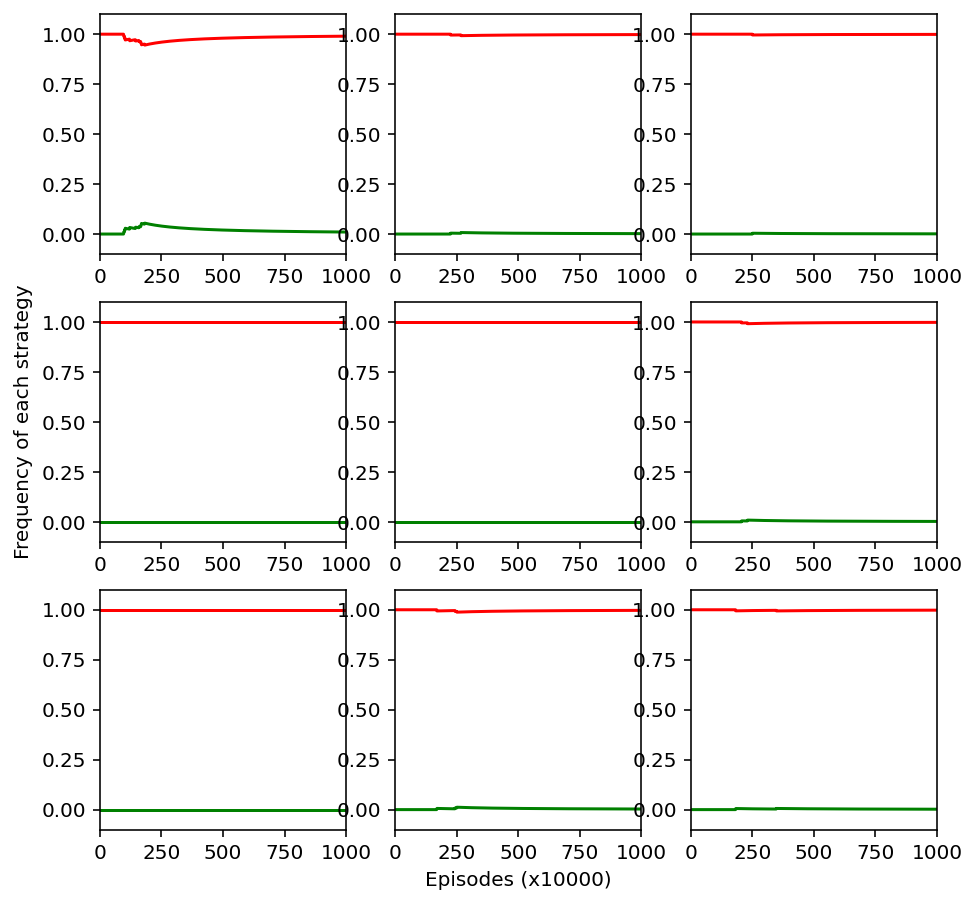}
    	\caption{$9$-headed DQN on Simple Artificial Chemistry}
        \label{fig:simple_chemistry_plot_DQN}
\end{figure}

\begin{figure}
        \hspace*{-0.5cm}
    	\includegraphics[height=0.9\linewidth]{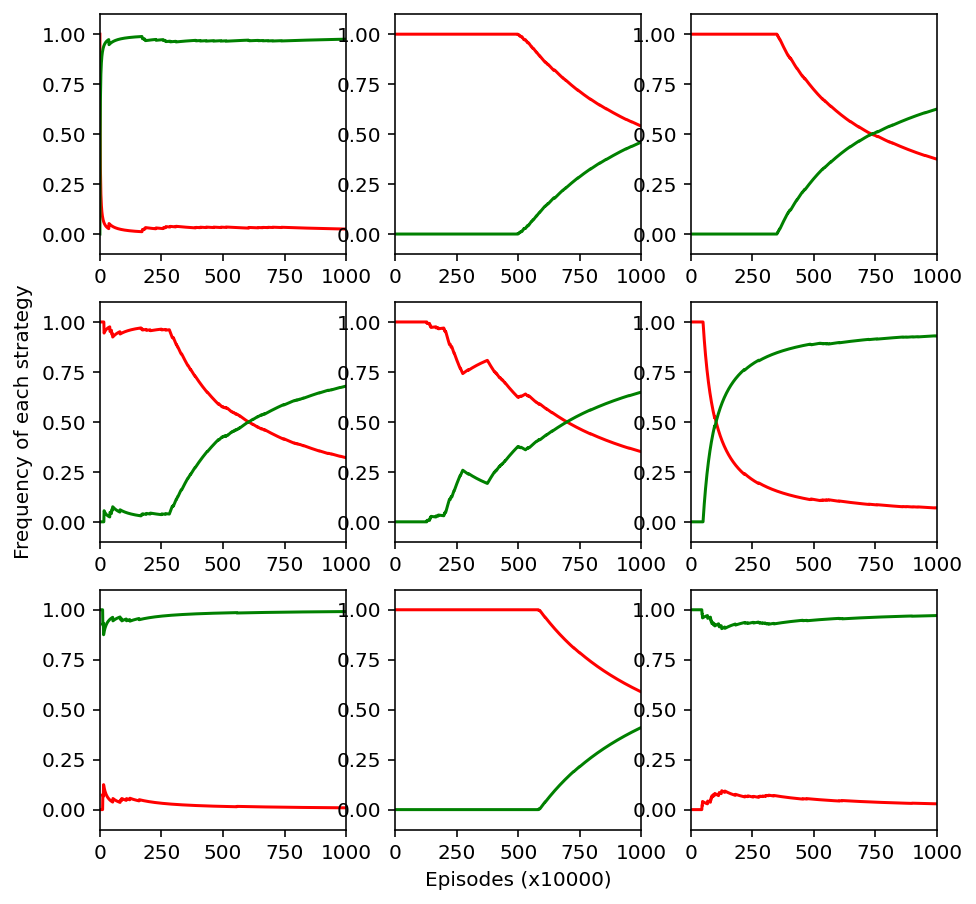}
    	\caption{$9$-headed DTE on Simple Artificial Chemistry}
        \label{fig:simple_chemistry_plot_DTE}
\end{figure}

Fig.~\ref{fig:simple_chemistry_plot_DQN} indeed shows that for DQN, all $9$ heads only utilize the red molecules while Fig.~\ref{fig:simple_chemistry_plot_DTE} shows that DTE early is having two heads going for the more rewarding green strategy while as learning goes on, eventually all heads learns to utilize the green pair reaction. Fig.~\ref{fig:simple_chemistry_best} shows a comparison of the best heads of DQN and DTE.

\begin{figure}
    	\includegraphics[width=8.0cm]{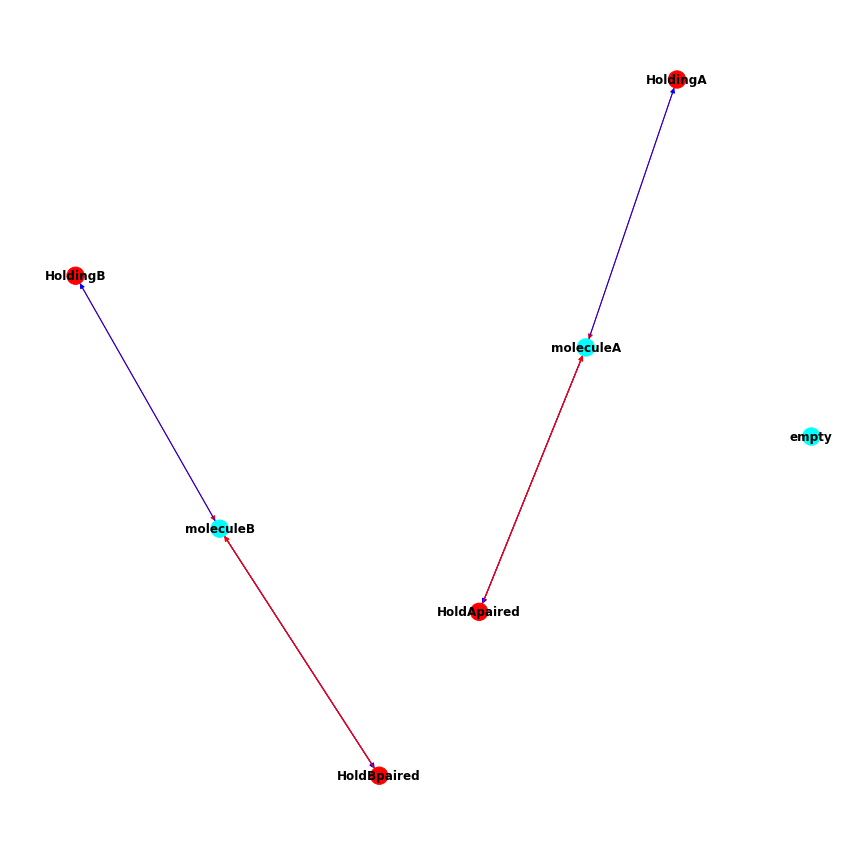}
    	\includegraphics[width=6.0cm]{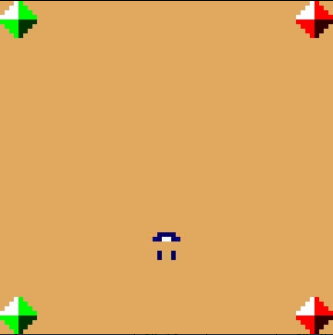}
    	\caption{Simple Artificial Chemistry. Reaction graph on the left and the map on the right. Left: The artificial chemistry we use represents the possible reactions with a directed multigraph~\citep{hagberg2008exploring}. The blue nodes are compounds and the red nodes are reactions. Arrows point from reactant compounds in to reaction nodes and arrows point out of reaction nodes toward product compounds. When double-headed, the same compound is both a reactant and a product and the two arrows overlap. See Fig.~\ref{fig:chemistry_metabolic} for a more complex example.}
        \label{fig:simple_chemistry_graph}
\end{figure}


\subsubsection{Metabolic cycles with distractors:}

Individuals benefit from different food generating cycles of reactions that rely on energy which dissipates over time if unused (red molecules). There are two possible autocatalytic reaction cycles, one cycle consists of molecules in three different shades of blue and the other cycle of molecules in three different shades of green. Both cycles require energy to continue. When they progress, they generates side products that comes in two types. If the two are brought together then that reaction generates new energy such that the cycles can continue, as well as a high reward for the agent. The population needs to keep both autocataytic cycles running in order to generate the side products for reward and also to make enough energy to sustain the system over time\footnote{see \textcolor{blue}{\url{https://youtu.be/FSStB7RpYis}} for an anonymized example video.}. 

\begin{figure}
    	\includegraphics[width=0.9\linewidth]{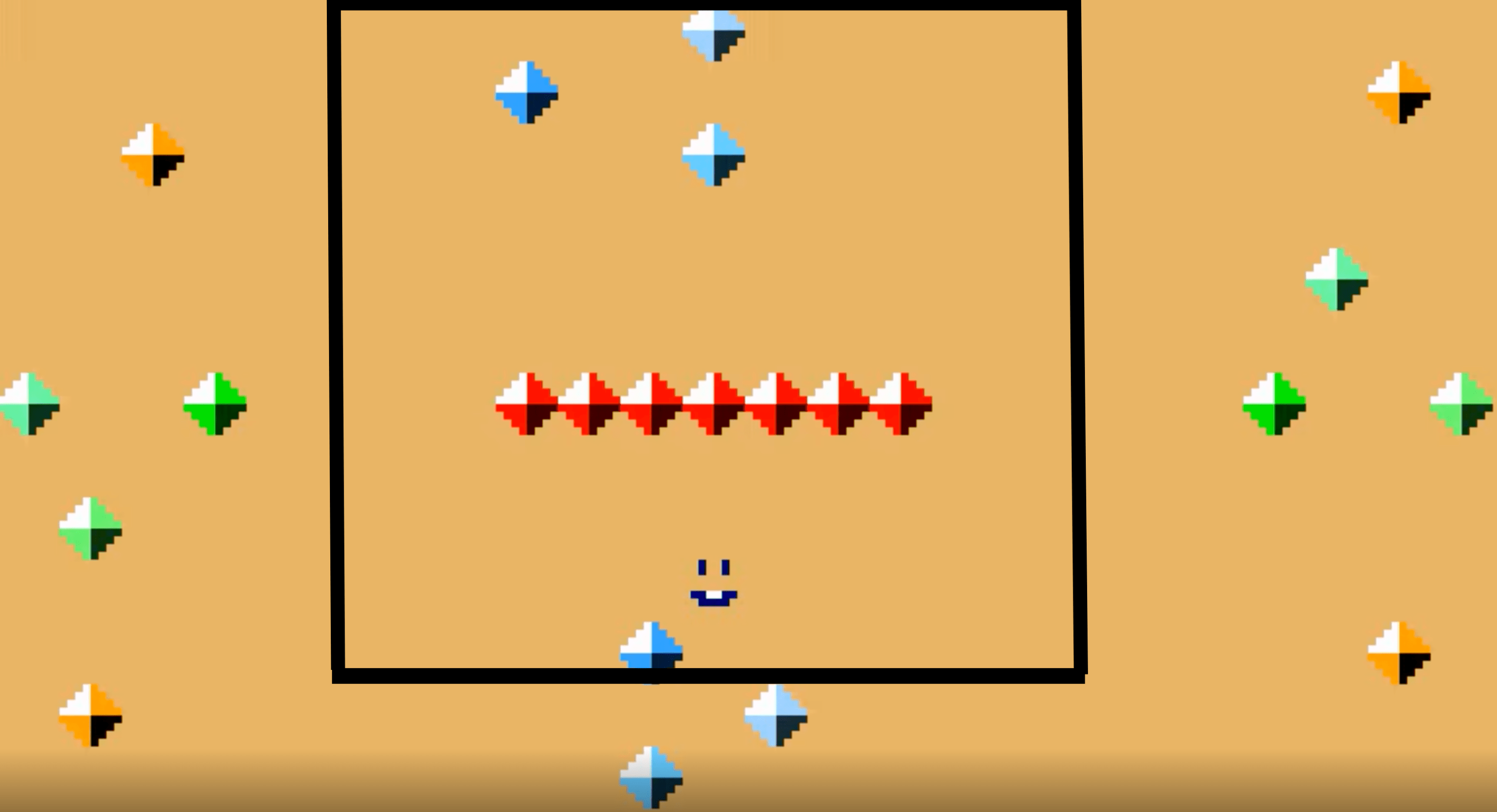}
    	\caption{Initial arrangement of molecules in Chemistry Metabolic Cycles with Distractors. The agent's partial viewing window is highlighted with a black rectangle. It is $11 \times 11$ sprites, each sprite being $8 \times 8$ pixels (making an $88 \times 88 \times 3$ RGB image observation).}
        \label{fig:chemistry_metabolic_pic}
\end{figure}

This task also contains one easier way to earn a small amount of rewarding and it is achieved by simply holding a molecule we call the distractor (orange in Fig.~\ref{fig:chemistry_metabolic_pic} where we can see $4$ of them, $1$ towards each corner) in its inventory\footnote{see \textcolor{blue}{\url{https://youtu.be/87NSFEPL7fk}} for an anonymized example video.}. This is the simplest rewarding strategy and represents a shallow local optima that agents consistently learns very early. The other molecules are involved in cycles where a subset of molecules are cyclically producing each other as long as there is energy available (red in Fig.~\ref{fig:chemistry_metabolic_pic} where we can see $7$ of them in the middle). The energy molecules are involved in a lower rate dissipation reaction, which will have all initial energy disappear long before the episode ends. There are two cycles, one involves three kinds of blue molecules (seen top and bottom in Fig.~\ref{fig:chemistry_metabolic_pic}) and one that involves three kinds of green molecules (seen left and right in Fig.~\ref{fig:chemistry_metabolic_pic}) that generates food that can be metabolized with a small reward but also a side product. The two cycles generates different side products and if one of each are brought together they participate in a rewarding reaction that also produces a new energy molecule. 
 
\begin{figure}
    	\includegraphics[width=\linewidth]{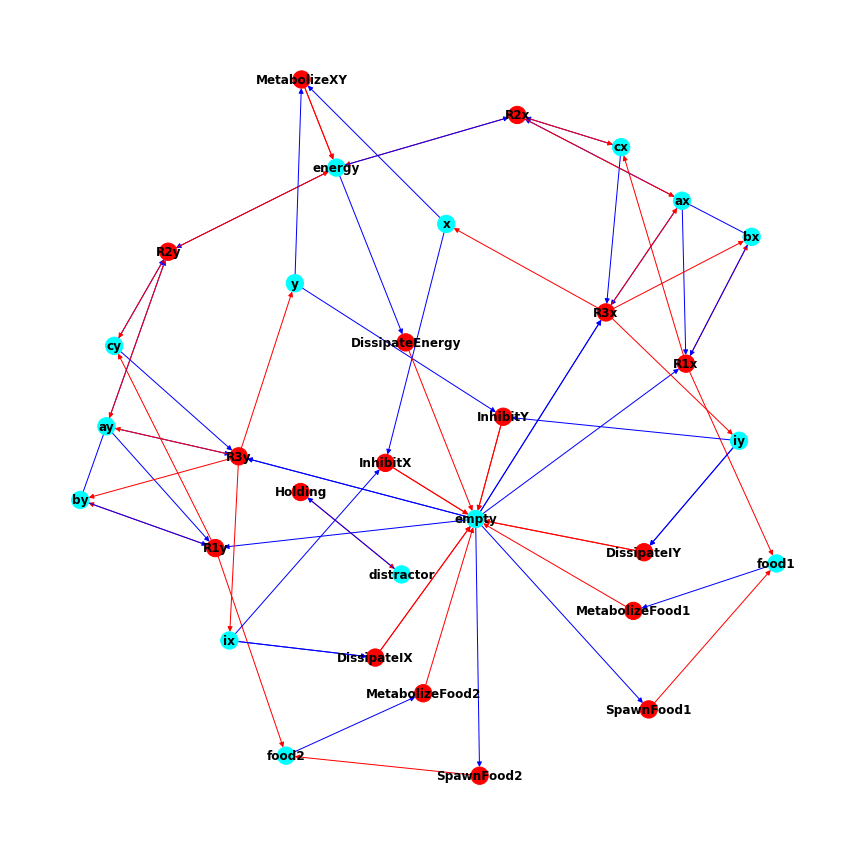}
     	\caption{Reaction Graph for molecules in Chemistry Metabolic Cycles with Distractors. See the caption of Fig.~\ref{fig:simple_chemistry_graph} for an explanation of how to interpret this graph representation.}
        \label{fig:chemistry_metabolic}
\end{figure}

\begin{figure}
		\includegraphics[width=1.0\linewidth]{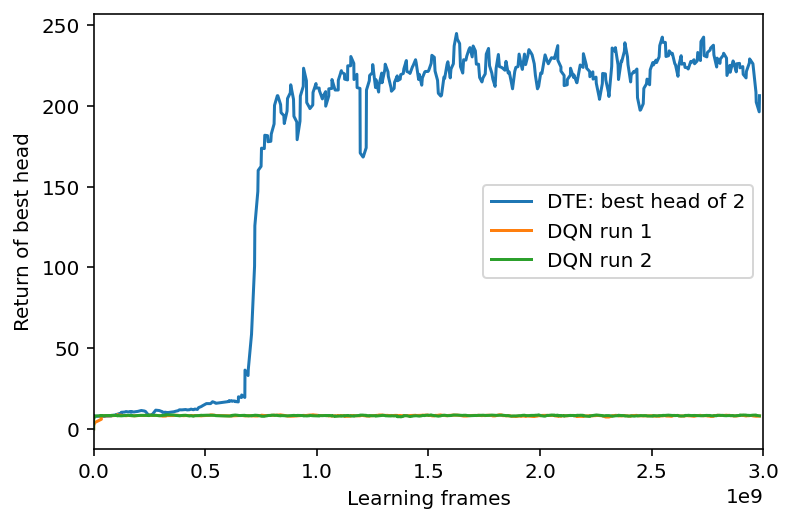}
    	\caption{Results for the Chemistry Metabolic Cycles with Distractors task.}
        \label{fig:chem_met}
\end{figure}
 
If an agents runs both blue and green cycles and brings the side products together it can all form a sustainable cycle that generates large amounts of reward for the agents. Agents receives reward for the reactions that metabolizes the two side products together, and a small reward for reactions that metabolizes the individual foods when they are in the inventory. The food metabolic reactions have much higher reactivity (probability of occurring) when in the inventory, and an agent is only rewarded when a rewarding reaction takes place in the inventory. 
 
We are going to display our approach's ability to find the harder niches despite having a very low $\varepsilon=0.01$, which the baselines will show is insufficient to leave the shallow optima and their performance quickly turns into a flat line. As this is a much more demanding task (more difficult, slower steps and longer episodes of length $1000$) then the others, having many heads leads to very long runs if each head is going to receive sufficient training episodes such that they could potentially learn the harder strategy well. We will, therefore, introduce one more feature to our approach and that is to have a fast learning heads and slow learning heads in the sense that they get sampled by a higher or lower probability and get many or fewer training episodes. We will here utilize the simplest implementation of that and let one head get $5$ times more training episodes than the others who are all equal. We will see that the faster head will depart from the distractor niche. Introducing asymmetry into the contenders for a niche to more quickly have some agent depart aligns well with our motivation. We consider taking this further in our future work, and train agents sequentially; first an agent that will learn to occupy the simplest niche and then the next finds it already inhabited with the value fully realized in the other agents value function.

As baseline here we use a standard single-agent DQN agent (with $\varepsilon=0.01$), i.e. DQN with one head in the terminology of this article, such that this head gets all the training episodes. As can be seen in Fig.~\ref{fig:chem_met}, this agent's episode returns always stays flat at the return for utilizing the distractor while DTE with two or nine heads is having its best head (the fast head) learning the more rewarding strategy of sustainably running all the cycles.

\section{Related Work} \label{sec:rel}

We here discuss related work that was not previously discussed, primarily algorithmic approaches related to diversity of a set of policies.
There is a technique in the evolutionary algorithms literature called fitness sharing \citep{goldberg1987genetic}. In fitness sharing, the total amount of fitness assigned to a particular individual in the population is decreased by the presence of other nearby individuals. This has an interpretation of overcrowding in niche space. This technique is particularly useful for problems where diversity is desirable, because it forces candidate solutions to be different. Because of this, it is extensively used in multi-objective optimization \citep{zhang2017survey, jaimes2017multi}. Our work could be viewed as using VDN-based techniques for implementing fitness sharing in RL agents.

There is a large literature utilizing mainly evolutionary methods to search for ``quality-diversity'', i.e. simultaneously satisfying high fitness and high diversity (e.g.~\citep{pugh2016QD}). Similarly in RL, there is work that uses diversity as a regularizer for reinforcement learning algorithms that aim to produce a set of policies \citep{Hong2018DivExpl, Masood2019DiversityInducingPG}. In particular, DOMINO is a continuous control algorithm that aims to find a set of diverse but all near-optimal policies \citep{Zahavy2022Domino}. The diversity objective was added in by the designer, not driven by the default reward. Our work however is based on a different underlying idea. Instead of aiming for a general form of diversity considered to be separate from the reward structure as the algorithms in this literature do, our proposed algorithm instead aims to discover all rewarding niches. 

The most important difference to most of the work discussed above, is that the final goal with DTE is to find more rewarding niches. It is not diversity for its own sake. In that sense it resembles in spirit the hierarchical RL literature on subgoal discovery. For instance feudal reinforcement learning \citep{dayan1992feudal, vezhnevets2017feudal} aims to find sub-policies that can then be selected when they are deemed most useful by a higher level ``manager'' agent and the option keyboard \citep{barreto2019option} seeks to learn options that can be arbitrarily combined with one another to synthesize combined skills.

The multi-agent approach that directly simulates competitive exclusion (e.g.~\citep{leibo2019malthusian}) can be very effective at niche identification in environments with the right characteristics. It requires a variable number of players to compete for a limited resources though, and this may not be practical for many problems. Also, when deployment will ultimately be in a single-agent setting there is always a risk that the learning would be too heavily influenced by the interactions with other players in the world itself and so would not ultimately be useful in single-player mode. Our approach on the other hand emphasizes niche exclusion specifically, not all kinds of competition. Also, the direct multi-agent method can be very demanding in the number of agents required to fill up a niche. In Malthusian RL, the Clamity task required $30$ (see \citep{leibo2019malthusian}). By contrast, the approach we described here allows for generalization between distractor stimuli (at least if they look similar to one another). So it could potentially get away with fewer sub-agents.

\section{Conclusion} \label{sec:conc}
We introduced the Diversity Through Exclusion (DTE) algorithm for niche identification in single-agent reinforcement learning, aiming to utilize the competitive exclusion principle to prevent premature convergence to poor local optima. We also introduced a chemistry inspired platform and showed how it can be used for creating tasks with multiple separate niches, something that we argued is lacking in many commonly used reinforcement learning tasks. We showed in both a simpler maze task as well as a more complex chemistry task, how our algorithm and its extra loss term enables the DTE agent to overcome less rewarding attractors and explore other niches in these tasks. The artificial chemistry platform and multi-agent enabled environments built with it are being contributed to the Melting Pot evaluation suite \citep{leibo2021scalable}. 

\bibliography{main}

\end{document}